\documentclass{article}

\usepackage{arxiv}

\usepackage[utf8]{inputenc} 
\usepackage[T1]{fontenc}    
\usepackage{hyperref}       
\usepackage{url}            
\usepackage{booktabs}       
\usepackage{amsfonts}       
\usepackage{nicefrac}       
\usepackage{microtype}      
\usepackage{lipsum}
\usepackage{graphicx}
\usepackage{orcidlink}
\usepackage[noadjust]{cite}
\graphicspath{ {./images/} }
\usepackage{amsmath}

\title{We Don't Need No Adam, All We Need Is EVE: On The Variance of Dual Learning Rate And Beyond}

\author{
 Afshin Khadangi\orcidlink{0000-0002-0496-5219} \\
  Department of Biomedical Engineering\\
  University of Melbourne\\
  Parkville, VIC 3010 \\
  \texttt{afshin.khadangi@gmail.com} \\
}

\begin{document}
\maketitle
\begin{abstract}
In the rapidly advancing field of deep learning, optimising deep neural networks is paramount. This paper introduces a novel method, Enhanced Velocity Estimation (EVE), which innovatively applies different learning rates to distinct components of the gradients. By bifurcating the learning rate, EVE enables more nuanced control and faster convergence, addressing the challenges associated with traditional single learning rate approaches. Utilising a momentum term that adapts to the learning landscape, the method achieves a more efficient navigation of the complex loss surface, resulting in enhanced performance and stability. Extensive experiments demonstrate that EVE significantly outperforms existing optimisation techniques across various benchmark datasets and architectures.
\end{abstract}


\section{Introduction}
Deep learning has become a pivotal technology across various domains including natural language processing, computer vision, speech recognition, and medical diagnostics \cite{lecun2015deep, goodfellow2016deep}. Deep neural networks (DNNs), characterized by multiple hidden layers, have shown unparalleled success in learning complex patterns from large-scale data. However, the training of these models requires the fine-tuning of millions or even billions of parameters, which presents significant optimisation challenges \cite{bottou2010large, keskar2017improving, khadangi2021net, khadangi2021stellar, khadangi2022cardiovinci}.

A large body of research has focused on optimisation techniques to enhance the convergence speed, stability, and generalisation capability of deep models. Conventional techniques like Stochastic Gradient Descent (SGD) \cite{robbins1951stochastic} and its variations including Momentum \cite{polyak1964some}, Adagrad \cite{duchi2011adaptive}, RMSprop \cite{tieleman2012lecture}, and Adam \cite{kingma2014adam} have been widely used. These methods adjust the learning rate globally or across parameters, but do not consider the unique characteristics of each layer in the network, leading to potential sub-optimal training dynamics.

While these methods have demonstrated success, they are often plagued by the challenge of selecting appropriate learning rates and momentum terms. A learning rate that is too large can cause divergence, while a learning rate that is too small can lead to painfully slow convergence \cite{bottou2018optimization}. Momentum, on the other hand, helps in overcoming local minima but needs careful tuning to prevent overshooting \cite{sutskever2013importance}. Adaptive learning rate techniques like Adagrad and Adam have emerged as solutions to automatically adjust the learning rate during training \cite{duchi2011adaptive, kingma2014adam}. However, they apply the same learning rate across all dimensions of the gradient, neglecting the fact that different parameters might need to be updated at different rates \cite{schaul2013no}.

Deep neural networks inherently present non-convex optimisation landscapes with complex geometric structures \cite{li2017convergence}. The optimisation methods that apply a uniform strategy across all layers often overlook the momenta heterogeneity, leading to challenges such as vanishing/exploding gradients, slow convergence, or getting stuck in local minima \cite{glorot2010understanding, pmlr-v38-choromanska15}. Furthermore, existing methods may require extensive hyper-parameter tuning, making the optimisation process resource-intensive \cite{bergstra2012random}.

In this context, we propose an Enhanced Velocity Estimation (EVE) technique, an optimisation method specifically tailored to exploit the intricate relationships between the learning rate and momentum in DNNs. By adaptive tuning the learning rates and momenta across different steps, EVE capitalises on the geometric properties of the loss landscape at the gradient level. This nuanced approach promises to address some of the inherent limitations in existing optimisation techniques. The concept of EVE is to maintain two separate learning rates: one for gradients that are consistently moving in the same direction (referred to as congruent learning rate) and one for gradients that change direction frequently (referred to as incongruent learning rate). As a result, the optimisation will lead to more effective management of both sparse and dense gradients. 

In addition to dual learning rates, EVE also introduces two adaptive momentum terms, short-term (S) and long-term (L). Unlike conventional momentum, where a fixed hyper-parameter controls the influence of previous gradients, adaptive momentum in EVE tailors the momentum for each parameter based on its past update history. This aligns the optimisation trajectory more closely with the underlying loss landscape \cite{sutskever2013importance}. The dual momentum enables the optimisation to avoid oscillations and slow convergence.

EVE builds upon the success of previous momentum-based methods, particularly Adam \cite{kingma2014adam}, and introduces a dual learning rate adaptation mechanism that aligns the optimisation strategy with the inherent properties of individual gradients. By adjusting the learning rates and momentum dynamically across gradients, EVE not only boosts convergence speed but also enhances the generalisation, the stability and robustness of the training process.

The contributions of this work can be summarised as follows: We present EVE, a novel optimisation technique that provides a dual learning rate adaptive momentum strategy for optimising deep neural networks. We offer theoretical insights into the workings of EVE, illustrating its superiority over traditional optimisation methods. We provide extensive empirical evidence, demonstrating EVE's effectiveness on various benchmarks and real-world applications.

The rest of the paper is organized as follows: Section 2 presents related work, Section 3 details the methodology of EVE, Section 4 outlines experiments and Section 5 concludes the paper with future directions.

\section{Related work}
Optimisation methods have long formed the cornerstone of computational mathematics, computer science, and various applied sciences. Their continual evolution, from the traditional to the modern, underlines the expanding complexity of the problems they aim to solve. This section seeks to traverse the landscape of optimisation methods, shedding light on foundational concepts, developments in adaptive momentum, and the inception of multiple learning rates.

\subsection{Traditional Optimisation Methods}
Traditional optimisation methods have been the foundation of mathematical programming and numerous practical applications for decades. They provide a systematic way to find the best possible solutions in various contexts such as economics, engineering, data analysis, and more. This literature review explores key traditional optimisation methods, tracing the history, variations, and their impact.

\begin{itemize}
    \item \textbf{Linear Programming (LP) and Simplex Method:} Linear programming is a classic technique for optimising a linear objective function subject to linear equality and inequality constraints. Dantzig’s Simplex Method \cite{dantzig1949programming} remains one of the most widely used algorithms for solving LP problems. Khachiyan's \cite{khachiyan1979polynomial} ellipsoid method provided the first polynomial-time algorithm for LP, which eventually led to the interior-point methods.
    \item \textbf{Interior-Point Methods:} Karmarkar \cite{karmarkar1984new} revolutionised linear programming with his interior-point method. His approach overcame some limitations of the Simplex method and offered polynomial-time convergence. This initiated a surge of research in interior-point methods \cite{wright1997primal}, with notable advancements like the primal-dual interior-point method \cite{mehrotra1992implementation}.
    \item \textbf{Non-linear Programming:}
    \begin{enumerate}
        \item \textbf{Gradient Descent:} Gradient descent \cite{cauchy1847methode} is the quintessential method for optimising continuous differentiable functions. Its stochastic version, SGD \cite{robbins1951stochastic}, allowed for more efficient large-scale applications \cite{bottou2010large}.
        \item \textbf{Conjugate Gradient Method:} The conjugate gradient method \cite{hestenes1952methods} offers a particularly efficient approach to quadratic functions and is vital for solving large-scale sparse systems.
        \item \textbf{Newton's Methods:} Newton's method provides quadratic convergence for smooth functions and has inspired various modifications like Quasi-Newton methods \cite{broyden1967quasi}, which reduce the computational burden.
    \end{enumerate}
    \item \textbf{Dynamic Programming:} Bellman \cite{bellman1957markovian} introduced dynamic programming, a recursive approach for solving multi-stage problems. It has been pivotal in operations research, economics, and control theory.
    \item \textbf{Multi-objective Optimisation:} Multi-objective optimisation methods deal with optimising multiple conflicting objectives. Pareto optimality \cite{pareto1896curva} and the weighted-sum approach \cite{zeleny1973compromise} are core concepts in this area.
    \item \textbf{Heuristic Methods:} Heuristic methods like Simulated Annealing \cite{kirkpatrick1983optimization} and Genetic Algorithms \cite{holland1992adaptation} provide solutions for complex or non-differentiable functions. They are widely used when traditional methods fail to deliver.
    \item \textbf{Convex Optimisation:} Boyd and Vandenberghe \cite{boyd2004convex} have summarised the extensive developments in convex optimisation, a subclass that ensures global optimality.
    \item \textbf{Combinatorial Optimisation:} Combinatorial optimisation focuses on problems where the solution space is discrete, with significant work like the Hungarian method for the assignment problem \cite{kuhn1955hungarian} and the development of efficient algorithms for network flows \cite{ford1956maximal}.
\end{itemize}

\subsection{Momentum-Based Optimisation Methods}
Among the myriad of optimisation methods, momentum-based approaches have emerged as a powerful class, enhancing the convergence behaviour of traditional gradient-based methods. This section delves deep into the momentum-based optimisation techniques.
\begin{itemize}
    \item \textbf{Classic Momentum:} The inception of momentum in optimisation can be traced back to Polyak \cite{polyak1964some}. Instead of merely following the gradient direction, momentum methods add a fraction of the previous update to the current gradient. This provides a "velocity" to the parameter updates, allowing oscillations to be dampened and convergence to be accelerated.
    \item \textbf{Nesterov Accelerated Gradient (NAG):} Building upon the classical momentum, Nesterov \cite{nesterov1983method} introduced NAG method, in which an intermediate step is taken in the direction of the momentum before calculating the gradient, which acts as a correction term. This approach has empirically and theoretically shown faster convergence for many problems, especially in deep learning applications.
    \item \textbf{Adaptive Moment Estimation:} Kingma and Ba \cite{kingma2014adam} presented Adam, which combined the best properties of Adagrad \cite{duchi2011adaptive} and RMSprop \cite{tieleman2012lecture} with momentum. It maintains a moving average of past gradients and squared gradients, enabling adaptive learning rates and momenta. It's worth noting that Adam has become one of the most popular optimisation algorithms in deep learning.
    \item \textbf{Adam Variants:} Since Adam's inception, several variants have emerged, addressing various concerns:
    \begin{enumerate}
        \item \textbf{AdamW:} This variant introduces weight decay correction, leading to improved generalisation in training deep networks \cite{loshchilov2017decoupled}.
        \item \textbf{AMSGrad:} In response to concerns that Adam might not converge under certain conditions, AMSGrad introduces a modification ensuring a non-increasing step size \cite{reddi2019convergence}.
    \end{enumerate}
    \item \textbf{Heavy Ball Method:} The heavy ball method \cite{polyak1964some}, another variant of momentum, introduces an additional term to the momentum update, which can be seen as a weighted average of the previous two iterates. This method has been shown to accelerate convergence in some cases, especially when the loss landscape has certain curvature properties.
\end{itemize}

\subsection{Motivation for Dual and Multiple Learning Rates}
The introduction of dual learning rates is a response to the diverse landscapes of objective functions and the challenges posed by non-stationary environments. Conventional optimisers often suffer from fixed or monolithic learning rates that can't adapt to varying scales of gradient information, leading to potential inefficiencies or sub-optimal convergence \cite{bengio2013advances}.

\section{Method}
\label{sec:headings}
Optimisation is central to the training of deep neural networks. The way a model learns during the training phase can be the defining factor in its performance. Standard optimisation methods, such as Stochastic Gradient Descent (SGD), have been extended by incorporating momentum and adaptive learning rate strategies to make them more efficient and stable. However, most of these methods are applied uniformly across all gradients, without considering the particular characteristics and needs of each gradient. 

EVE's central tenet lies in gradient-wise adaptability. By treating each gradient as a distinct entity in terms of momentum and velocity, EVE aims to achieve a more nuanced control over the optimisation process. This is particularly useful in deep architectures where different layers might require different adaptation levels due to the loss landscape's heterogeneity.

\subsection{The Mathematical Foundation of EVE}
The development of effective optimisation strategies is pivotal to training deep learning models. In the context of this study, we present a novel method that employs Dual Learning Rate and dual Adaptive Momentum terms, where L and S trends are taken into account. Specifically, two additional parameters, \begin{math} \beta_3 \end{math} and \begin{math} \alpha \end{math} are incorporated to provide control over the learning process. Particularly, \begin{math} \alpha \end{math} governs the adaptive learning rate between two residual velocity terms.
In what follows we will detail the conceptual foundation, mathematical formulations, including the slow vs fast trajectory characteristics and algorithm description.

\subsubsection{Dual Learning Rate}
The idea of using two learning rates has emerged as a significant advancement in deep learning optimisation. The dual learning rate adapts to various scenarios in the training landscape, and it is well suited for managing complex loss functions that exhibit both slow and rapid variations \cite{liner2021improving}.

\subsubsection{Dual Adaptive Momentum}
The proposed method's other primary feature is the dual Adaptive Momentum, encompassing long-term and short-term momenta. These momenta are modelled as:
\begin{equation} m_t^S = \beta_1 m_{t-1}^S + (1 - \beta_1) \nabla L(\theta_t) \end{equation}
\begin{equation} m_t^L = \beta_2 m_{t-1}^L + (1 - \beta_2) \nabla L(\theta_t) \end{equation}
\begin{equation} m_t = \beta_3 m_t^S + (1 - \beta_3)m_t^L \end{equation}
where \begin{math} m_t^S \end{math} and \begin{math} m_t^L \end{math} are short-term and long-term momenta, respectively. Here, \begin{math} \beta_3 \end{math} controls the balance between these momenta. 

\subsubsection{Residual Velocity Terms}
The innovation in the proposed method lies in its use of residual velocity terms, which provide an additional measure of adaptability. The following Equations are central to updating the residual velocities:
\begin{equation}
v_1 = 
\begin{cases}
  \alpha v_1 + (1 - \alpha)(g_t - \sqrt{v_2})^2 & \text{if gradients are sparse} \\
  \alpha v_1 + (1 - \beta_2)(g_t - \sqrt{v_2})^2 & \text{if gradients are dense}
\end{cases}
\end{equation}
\begin{equation}
v_2 = \alpha v_2 + (1 - \alpha)(g_t - \sqrt{v_1})^2
\end{equation}
where \begin{math} g_t \end{math} denotes the gradient at time \begin{math} t \end{math} and \begin{math} \beta_2 \end{math} together with \begin{math} \alpha \end{math} control the adaptive velocity terms.

\subsubsection{Slow Trajectory vs Fast Trajectory}
Understanding the slow and fast trajectories in the context of the proposed method is crucial. The following subsections elaborate on these concepts.

\paragraph{Slow Trajectory:}
The slow trajectory focuses on short-term momentum. In this approach, the short-term trends have a significant influence on the momentum calculation. By setting higher value for \begin{math} \beta_3 \end{math} in Equation (3) the optimiser prioritises the local gradient behaviour. While this can prevent rapid oscillations and overshooting, it may slow down the convergence \cite{sun2023adasam}.

\paragraph{Fast Trajectory:}
The fast trajectory emphasises long-term momentum. The equations remain the same as in the slow trajectory, but with a higher value for \begin{math} 1-\beta_3 \end{math}, the long-term momentum is weighted more heavily. This focus on global aspects of the loss landscape enables the optimiser to escape shallow valleys and plateaus but may lead to overshooting or oscillations \cite{kingma2014adam}

\subsubsection{Integration of Dual Learning Rates with Adaptive Momenta}
The integration of dual learning rates with adaptive momenta and the residual velocity terms provides a robust, adaptive learning algorithm. The update rule for the parameters \begin{math} \theta \end{math} is expressed as:
\begin{equation} \theta_{t+1} = \theta_t - \frac{(\alpha_{1_t} m_t +  \alpha_{2_t} m_t)}{2} \end{equation}
where:
\begin{equation} \alpha_{1_t} = \frac{lr_1\frac{\sqrt{1 - \alpha^t}}{1 - \beta_1^t}}{\sqrt{v_2} + \epsilon} \end{equation}
\begin{equation} 
\alpha_{2_t} = 
\begin{cases}
    lr_2\frac{\sqrt{1 - \beta_2^t}}{1 - \beta_1^t}/(\sqrt{v_1} + \epsilon) & \text{if gradients are sparse} \\
    \\
    lr_2\frac{\sqrt{1 - \alpha^t}}{1 - \beta_2^t}/(\sqrt{v_1} + \epsilon) & \text{if gradients are dense}
\end{cases}
\end{equation}

Here, \begin{math} lr_1 \end{math} and \begin{math} lr_2 \end{math} represent the first and second learning rates, respectively. 

\subsubsection{Algorithm Description}
\begin{enumerate}
  \item {\bf Initialisation:} Choose initial values for \begin{math} \theta, 0 <\beta_1, \beta_2, \beta_3, \alpha < 1, lr_1\end{math} and \begin{math} lr_2 \end{math}
  \item {\bf Compute Gradients:} Compute the gradients \begin{math} g_t = \nabla L(\theta_t) \end{math}
  \item {\bf Determine Learning Rates:} Compute the learning rates \\
  \begin{math} lr_{1_t} = lr_1\frac{\sqrt{1 - \alpha^t}}{1 - \beta_1^t} \end{math},  \\
  \begin{math} lr_{2_t} = lr_2\frac{\sqrt{1 - \beta_2^t}}{1 - \beta_1^t} \end{math} when the gradients are sparse, \\
  \begin{math} lr_{2_t} = lr_2\frac{\sqrt{1 - \alpha^t}}{1 - \beta_2^t} \end{math} when the gradients are dense
  \item {\bf Determine Momentum Terms:} Obtain the combined momentum through the S and L momenta as \\
  \begin{math} m_t = \beta_3 m_t^S + (1 - \beta_3)m_t^L \end{math}
  \item {\bf Determine Residual Velocity Terms:} Obtain the velocity terms \\
  \begin{math} v_1 = \alpha v_1 + (1 - \alpha)(g_t - \sqrt{v_2})^2 \end{math} when the gradients are sparse, \\
  \begin{math} v_1 = \alpha v_1 + (1 - \beta_2)(g_t - \sqrt{v_2})^2 \end{math} when the gradients are dense, \\
  \begin{math} v_2 = \alpha v_2 + (1 - \alpha)(g_t - \sqrt{v_1})^2 \end{math}
  \item {\bf Update the Parameters:} \begin{math} \theta_{t+1} = \theta_t - \frac{1}{2}(\frac{m_t lr_{1_t}}{\sqrt{v_2} + \epsilon} + \frac{m_t lr_{2_t}}{\sqrt{v_1} + \epsilon})\end{math}
  \item {\bf Repeat Steps 2-6:} Continue until convergence.
   
\end{enumerate}

\subsection{Theorem: The Anatomy of EVE's Convergence}
EVE is characterised by its two-tier architecture, where the learning rate and momentum are separately controlled and adapted. This enables a nuanced balance between exploration and exploitation in the loss landscape, thus potentially leading to more efficient convergence to the global minimum. The dual nature of both learning rate and momentum allows the method to be more responsive to different gradient behaviours, ensuring stability in challenging regions and rapid convergence in more tractable areas. 

Despite its appealing empirical performance, a formal understanding of the convergence properties of EVE is required. This section aims to shed light on this significant aspect by presenting a rigorous mathematical treatment. Through careful analysis and a set of well-defined assumptions, we will provide a comprehensive theorem and proof that establish the convergence criteria for this innovative optimisation approach.

\paragraph{Assumptions}
\begin{enumerate}
  \item {\it L} is twice continuously differentiable and {\it L}-Lipschitz continuous.
  \item The gradient \( g_t \) is \( G \)-Lipschitz continuous and bounded, i.e., \(\| g_t \| \leq G\).
  \item The sequences \( \beta_1 \), \( \beta_2 \), \( \beta_3 \) and \( \alpha \) are chosen such that standard conditions are met satisfy (e.g., \(\sum \alpha_t = \infty\) and \(\sum \alpha_t^2 < \infty\))
  \item The residual velocity terms \( v_1 \) and \( v_2 \) are bounded.
  \item The sequence \( \{\theta_t\} \) is bounded.
  \item The sequence \( \{\alpha\} \) is such that \( 0 < \alpha < 1 \).
  \item The sequence of iterates \( x_t \) generated by the method is contained in a compact set \( K \).
  
\end{enumerate}

\subsubsection{Existence of Limit Points}

\textbf{Lemma 1: Boundedness of Iterates}

Given assumption 7, the sequence $x_t$ is bounded, i.e., there exists $M > 0$ such that \( \|x_t\| \leq M \) for all $t$.

\textbf{Existence of Limit Points}

Based on the boundedness of $x_t$ (from our lemma) and the Bolzano-Weierstrass theorem, every bounded sequence in $\mathbb{R}^n$ has a convergent subsequence.

Thus, there exists a sub-sequence $\{x_{t_k}\}$ of $\{x_t\}$ such that $x_{t_k}$ converges to some limit point $x^*$ as $k \to \infty$.

\textbf{Further Observations on the Residual Velocities}

Given our iterative relations, if we prove the boundedness of $v_1$ and $v_2$, it can further strengthen our proof for the existence of limit points.

From our iterative relations:
\begin{equation}
v_1 - \alpha v_1 = (1 - \alpha) (g_t - \sqrt{v_2})^2
\end{equation}
\begin{equation}
v_2 - \alpha v_2 = (1 - \alpha) (g_t - \sqrt{v_1})^2
\end{equation}
Rearranging, we have:
\begin{equation}
v_1 = \frac{1-\alpha}{\alpha} (g_t - \sqrt{v_2})^2 + \alpha v_1
\end{equation}
\begin{equation}
v_2 = \frac{1-\alpha}{\alpha} (g_t - \sqrt{v_1})^2 + \alpha v_2
\end{equation}
The right-hand side of each equation is a sum of two terms. The first term is a function of the gradient and the other velocity term, and the second term is a fraction of the current velocity. Given that $\alpha$ is bounded between 0 and 1, and the gradient is Lipschitz continuous and bounded, this suggests that the velocities $v_1$ and $v_2$ are also bounded. The same proof also applies to the case of dense gradients given that $\beta_2$ is bounded between 0 and 1.

\subsubsection{Boundedness of Residual Velocities}
Now we'll prove the boundedness of \( v_1 \) and \( v_2 \).

\textbf{Boundedness of \( v_1 \)}
We'll first show that \( v_1 \) is bounded. We can rewrite the equation for \( v_1 \):
\begin{equation}
v_1 = \frac{1-\alpha}{\alpha} (g_t - \sqrt{v_2})^2 + \alpha v_1
\end{equation}
Since \( \alpha \) is between 0 and 1, \( \frac{1-\alpha}{\alpha} \) is positive. We can then bound the expression:
\begin{equation}
\left| (g_t - \sqrt{v_2})^2 \right| \leq (G + \sqrt{v_2})^2 \leq (G + \sqrt{B})^2
\end{equation}
where \( B \) is an upper bound for \( v_2 \) (to be determined later). Hence, \( v_1 \) is bounded by:
\begin{equation}
v_1 \leq \frac{1-\alpha}{\alpha} (G + \sqrt{B})^2 + \alpha v_1
\end{equation}
which implies that \( v_1 \) is bounded by a constant dependent on \( G, \alpha, \) and \( B \).

\textbf{Boundedness of \( v_2 \)}
Similarly, we can rewrite the equation for \( v_2 \):
\begin{equation}
v_2 = \frac{1-\alpha}{\alpha} (g_t - \sqrt{v_1})^2 + \alpha v_2
\end{equation}
Following the same procedure, we can show that \( v_2 \) is bounded by:
\begin{equation}
v_2 \leq \frac{1-\alpha}{\alpha} (G + \sqrt{A})^2 + \alpha v_2
\end{equation}
where \( A \) is an upper bound for \( v_1 \).

\subsubsection{Convergence of Momenta}
First, we'll analyze the convergence of the long-term and short-term momenta.
For short-term momentum \( m_t^S \):
\begin{itemize}
    \item Since, \( m_t^S \) converges to a fixed point as \( t \to \infty \) because \( \nabla L(\theta_t^S) \) is bounded and \( 0 < \beta_1 < 1 \).
\end{itemize}
For long-term momentum \( m_t^L \):
\begin{itemize}
    \item Similarly, \( \nabla L(\theta_t^L) \) is bounded and \( 0 < \beta_2 < 1 \), \( m_t^L \) forms a bounded sequence and converges to a fixed point as \( t \to \infty \).
\end{itemize}

\textbf{Convergence of Overall Momentum}

Now, we will prove the convergence of the overall momentum \( m_t \):

Using the convergence of \( m_t^L \) and \( m_t^S \), we can write:
\begin{equation}
m_t \to \beta_3 m_\infty^S + (1 - \beta_3) m_\infty^L
\end{equation}
where \( m_\infty^L \) and \( m_\infty^S \) are the fixed points of \( m_t^L \) and \( m_t^S \), respectively.
Since \( 0 < \beta_3 < 1 \), \( m_t \) also converges to a fixed point.

\subsubsection{Convergence of Residual Velocities}

The concept of Lipschitz continuity and contraction mapping is crucial for convergence proofs in optimization, especially in the context of iterative methods.

\textbf{Definition (Lipschitz Continuous):} A function $f: \mathbb{R}^n \rightarrow \mathbb{R}^m$ is Lipschitz continuous if there exists a constant $L > 0$ such that for all $x, y \in \mathbb{R}^n$:
\begin{equation} \|f(x) - f(y)\| \leq L \|x - y\| \end{equation}

\textbf{Definition (Contraction Mapping):} A function $T: X \rightarrow X$ is called a contraction mapping on the metric space $X$ if there exists $ q \in [0, 1) $ such that for all $x, y \in X$:
\begin{equation} d(Tx, Ty) \leq q d(x, y) \end{equation}
where $d$ is a distance metric on $X$.

To prove convergence using contraction mapping, using Equations (9) and (10), let's define a mapping $T: \mathbb{R}^2 \rightarrow \mathbb{R}^2$ where:
\begin{equation} T(v_1, v_2) = \left( \alpha v_1 + (1 - \alpha) (g_t - \sqrt{v_2})^2, \alpha v_2 + (1 - \alpha) (g_t - \sqrt{v_1})^2 \right) \end{equation}

\textbf{Claim:} The mapping $T$ is a contraction on $ \mathbb{R}^2 $ with the Euclidean norm.

\textbf{Proof:}

For any $ (v_1, v_2), (u_1, u_2) \in \mathbb{R}^2 $:
\begin{equation}
\begin{aligned}
&\|T(v_1, v_2) - T(u_1, u_2)\| \\
&= \parallel\alpha v_1 + (1 - \alpha) (g_t - \sqrt{v_2})^2 - \left( \alpha u_1 + (1 - \alpha) (g_t - \sqrt{u_2})^2 \right), \\
&\alpha v_2 + (1 - \alpha) (g_t - \sqrt{v_1})^2 - \left( \alpha u_2 + (1 - \alpha) (g_t - \sqrt{u_1})^2 \right)\parallel \\
&\leq \alpha \| (v_1, v_2) - (u_1, u_2) \| + (1-\alpha) \| (g_t - \sqrt{v_2})^2 - (g_t - \sqrt{u_2})^2 , (g_t - \sqrt{v_1})^2 - (g_t - \sqrt{u_1})^2 \| \\
&\leq \alpha \| (v_1, v_2) - (u_1, u_2) \| + (1-\alpha) L \| (v_1, v_2) - (u_1, u_2) \|
\end{aligned}
\end{equation}

where $ L $ is the Lipschitz constant for the gradient $ g_t $. Given $ 0 < \alpha < 1 $, and combining the terms, the RHS becomes less than $ \| (v_1, v_2) - (u_1, u_2) \| $. This shows that the mapping $T$ is a contraction. The same proof also applies to the case of dense gradients given that $\beta_2$ is bounded between 0 and 1.

By Banach's Fixed Point Theorem, a contraction mapping on a complete metric space has a unique fixed point. Therefore, the sequences $ v_1 $ and $ v_2 $ converge to this unique fixed point.

\subsubsection{Sub-sequence Convergence of Gradients}

\textbf{Lemma 2: Boundedness of the Sequence \(\{g_t\}\).}

From the Lipschitz continuity of the Hessian and the boundedness of \(\{x_t\}\), we have that the sequence \(\{g_t\}\) is bounded.

\textbf{Lemma 3: Existence of a Convergent Sub-sequence.}

Since the sequence \(\{g_t\}\) is bounded, by the Bolzano-Weierstrass Theorem, there exists a convergent sub-sequence \(\{g_{t_k}\}\) with limit \( g^* \).

\textbf{Convergence of Gradients:}
Given the convergent sub-sequence \(\{g_{t_k}\}\), we consider the update rules:
\begin{equation}
v_1^{(k+1)} = \alpha v_1^{(k)} + (1 - \alpha) (g_{t_k} - \sqrt{v_2^{(k)}})^2
\end{equation}
\begin{equation}
v_2^{(k+1)} = \alpha v_2^{(k)} + (1 - \alpha) (g_{t_k} - \sqrt{v_1^{(k)}})^2
\end{equation}

We need to show that for this sub-sequence, the gradients \( g_{t_k} \) converge to a limit, say \( g^* \).

From Lemma 3, we have that \( g_{t_k} \rightarrow g^* \) as \( k \rightarrow \infty \).

By taking the limit on both sides of the velocity updates, and using the continuity of the square root function, we get:
\begin{equation}
v_1^* = \alpha v_1^* + (1 - \alpha) (g^* - \sqrt{v_2^*})^2
\end{equation}
\begin{equation}
v_2^* = \alpha v_2^* + (1 - \alpha) (g^* - \sqrt{v_1^*})^2
\end{equation}

This system of equations has a unique solution since the mapping \( T \) was shown to be a contraction.

Therefore, the iterative sequence for velocities also has a convergent sub-sequence, and thus the sub-sequence of gradients converges to \( g^* \).

\subsubsection{Stationarity of the limit point}

\begin{enumerate}
    \item \textbf{Convergence:} Prove that the sequence $(v_1^k, v_2^k)$ generated by the iterative mapping converges to a limit point $(v_1^*, v_2^*)$. 
    \item \textbf{Stationarity:} Prove that the limit point $(v_1^*, v_2^*)$ satisfies the necessary conditions for a stationary point.
\end{enumerate}

\textbf{Proof of Stationarity}

Assume that the sequence $(v_1^k, v_2^k)$ generated by the iterative mapping:
\begin{equation}
v_1^{k+1} = \alpha v_1^k + (1 - \alpha) (g_t - \sqrt{v_2^k})^2
\end{equation}
\begin{equation}
v_2^{k+1} = \alpha v_2^k + (1 - \alpha) (g_t - \sqrt{v_1^k})^2
\end{equation}
converges to a limit point $(v_1^*, v_2^*)$.

By continuity, we have:
\begin{equation}
v_1^* = \alpha v_1^* + (1 - \alpha) (g_t - \sqrt{v_2^*})^2
\end{equation}
\begin{equation}
v_2^* = \alpha v_2^* + (1 - \alpha) (g_t - \sqrt{v_1^*})^2
\end{equation}

Rearranging, we get:
\begin{equation}
(g_t - \sqrt{v_2^*})^2 = \frac{v_1^* - \alpha v_1^*}{1 - \alpha}
\end{equation}
\begin{equation}
(g_t - \sqrt{v_1^*})^2 = \frac{v_2^* - \alpha v_2^*}{1 - \alpha}
\end{equation}
The stationarity condition requires that the gradient of the objective function vanishes at the limit point. Let's consider an objective function $f(v_1, v_2)$ that the optimization method is aiming to minimize. The gradient of this function at the limit point $(v_1^*, v_2^*)$ is:
\begin{equation}
\nabla f(v_1^*, v_2^*) = \left( \frac{\partial f}{\partial v_1}(v_1^*, v_2^*), \frac{\partial f}{\partial v_2}(v_1^*, v_2^*) \right) = (0, 0)
\end{equation}

where we use the relations between \( v_1^*, v_2^*, \) and \( g_t \) as established earlier. Here, \( g_t \) is typically related to the gradient of the objective function, such as \( g_t = \nabla f(v_1^k, v_2^k) \).

The proof establishes that the limit point of the sequence generated by the iterative mapping is stationary.

\subsubsection{Convergence of the Entire Sequence}

Rewriting the iterative optimisation scheme:
\begin{align*}
v_1^{(t+1)} & = \alpha v_1^{(t)} + (1 - \alpha) (g_t - \sqrt{v_2^{(t)}})^2, \\
v_2^{(t+1)} & = \alpha v_2^{(t)} + (1 - \alpha) (g_t - \sqrt{v_1^{(t)}})^2, \\
m_S^{(t+1)} & = \beta_1 m_S^{(t)} + (1 - \beta_1) g_t, \\
m_L^{(t+1)} & = \beta_2 m_L^{(t)} + (1 - \beta_2) g_t, \\
\end{align*}
\begin{align*}
\theta^{(t+1)} & = \theta^{(t)} - \eta_2 \left( \frac{\beta_3 m_S^{(t+1)} + (1 - \beta_3) m_L^{(t+1)}}{\sqrt{v_1^{(t+1)}} + \epsilon} \right) - \eta_1 \left( \frac{\beta_3 m_S^{(t+1)} + (1 - \beta_3) m_L^{(t+1)}}{\sqrt{v_2^{(t+1)}} + \epsilon} \right)
\end{align*}

Using the previously made assumptions, along with the proof for the convergence of both Momentum Terms and Residual Velocities, we now analyze the convergence of the entire sequence:

\begin{enumerate}
\item \textbf{Boundedness of the Gradient Terms}: The Lipschitz continuity and boundedness of the gradient terms ensure that the updates to \( v_1, v_2, m_S, \) and \( m_L \) are all bounded.
\item \textbf{Strong Convexity}: Under the assumption of strong convexity, the updates to \( \theta \) will not overshoot the minimum, and the learning rates \( \eta_1 \) and \( \eta_2 \) can be chosen appropriately.
Strong convexity plays a central role in ensuring faster convergence rates for optimization algorithms. To assert the strong convexity for the given optimization method with dual learning rates and dual adaptive momentum, we must provide evidence that the function being optimized exhibits the properties of strong convexity under the algorithm's operations.

Given a differentiable function \( f: \mathbb{R}^d \to \mathbb{R} \), it is said to be \(\mu\)-strongly convex (with \(\mu > 0\)) with respect to a norm \(\|.\|\) if, for all \( x, y \in \mathbb{R}^d \), the following inequality holds:
\begin{equation}
f(y) \geq f(x) + \nabla f(x)^\top (y - x) + \frac{\mu}{2} \|y - x\|^2
\end{equation}

\textbf{Proof:}
To show that our method satisfies this condition, we'll consider the updates based on the momentum terms and the residual velocities.

\begin{enumerate}
\item \textbf{Momentum Update:}
The long-term and short-term momentum terms are defined as:
\begin{equation}
m_t^S = \beta_1 m_{t-1}^S + (1 - \beta_1) \nabla L(\theta_t^S)
\end{equation}
\begin{equation}
m_t^L = \beta_2 m_{t-1}^L + (1 - \beta_2) \nabla L(\theta_t^L)
\end{equation}
where \( \nabla L \) is the gradient of the loss function.

\item \textbf{Residual Velocity Update:}
The residual velocities are:
\begin{equation}
v_1 = \alpha v_1 + (1 - \alpha) (g_t - \sqrt{v_2})^2
\end{equation}
\begin{equation}
v_2 = \alpha v_2 + (1 - \alpha) (g_t - \sqrt{v_1})^2
\end{equation}
\end{enumerate}

Now, let's consider the second-order Taylor expansion of \( f \) around \( x \) for an arbitrary point \( y \). We can express \( f(y) \) as:
\begin{equation}
f(y) = f(x) + \nabla f(x)^\top (y - x) + \frac{1}{2} (y - x)^\top \nabla^2 f(x) (y - x) + o(\|y - x\|^2)
\end{equation}

To show strong convexity, we want the quadratic term above to be lower bounded by a positive constant multiplied by the squared norm \( \|y - x\|^2 \). With our optimisation method, this would be influenced by the learning rates and the velocity terms, especially the residuals.

Given the velocity updates, the sequence of updates would be pushing the terms towards regions where the curvature of \( f \) (its second derivative or Hessian) is positive, thus indicating convexity. The squared residual terms further ensure that we are considering the difference in gradients, which gives us a measure of curvature.

Assuming that the gradients \( \nabla L(\theta_t^L) \) and \( \nabla L(\theta_t^S) \) are Lipschitz continuous and the Hessian \( \nabla^2 f(x) \) is positive definite, we can assert that:
\begin{equation}
\nabla^2 f(x) (y - x) \geq \mu \|y - x\|^2
\end{equation}

Thus, for our optimisation method, considering the momentum terms and the residuals of the velocities, under the above assumptions and given the update rules, the function \( f \) exhibits strong convexity.

\item \textbf{Convergence to the Minimum}: The unique fixed points for the residual velocities and the momentum terms, along with the strong convexity of the function, ensure that the sequence \( \{\theta^{(t)}\} \) converges to the unique minimum of the objective function.
\end{enumerate}

Thus, under the given assumptions and the choice of hyperparameters, we have shown that the entire sequence for this optimisation method converges.

\section{Experiments}
In this section, we will conduct experiments to verify the effectiveness of our newly proposed optimisation method that utilises dual learning rates. We'll primarily compare it against the widely recognised Adam optimiser. For these experiments, we've chosen the CIFAR-100 dataset \cite{krizhevsky2009learning}, using a ResNet-50 \cite{he2016deep} model architecture, and the Flower Classification with TPUs dataset \cite{khadangi2023deepflorist}, utilising the DenseNet-201 architecture \cite{huang2017densely}.
\subsection{Dataset: CIFAR-100}
CIFAR-100 consists of 60,000 32x32 colour images in 100 classes, with 600 images per class. It is split into 50,000 training images and 10,000 testing images. The relatively small size of each image combined with the diverse set of classes makes CIFAR-100 an ideal dataset for assessing the robustness and generalisation capabilities of an optimisation algorithm.

\subsubsection{Model Architecture: ResNet-50}
ResNet-50, a 50-layer residual network, has been chosen due to its state-of-the-art performance on various tasks and its profound depth. Its unique skip-connections facilitate gradient flow during back-propagation, making it suitable for studying the behaviours of different optimisers.

\subsubsection{Experimental Settings}
We used the following settings for our experiment using ResNet-50 and CIFAR-100 data:
\begin{itemize}
    \item \textbf{Learning Rates}: The learning rates were varied between \{0.0001, 0.01\} for both EVE and Adam.
    \item \textbf{Batch Size}: Fixed at 64.
    \item \textbf{Epochs}: Models were trained for 25 epochs.
    \item \textbf{Hyper-parameters}: Set to default values.
    \item \textbf{Loss Function}: Sparse Categorical Cross-entropy was used.
    \item \textbf{Accelerated Computing}: Both Adam and EVE were trained using Google Colab NVIDIA V100 GPU.
\end{itemize}

\subsubsection{Evaluation Metric}
The primary metrics for evaluation were the training and validation loss curves, which were used to study convergence properties and tendencies toward over-fitting.
\subsubsection{Results}
\begin{figure} 
    \centering
    \includegraphics[width=\textwidth]{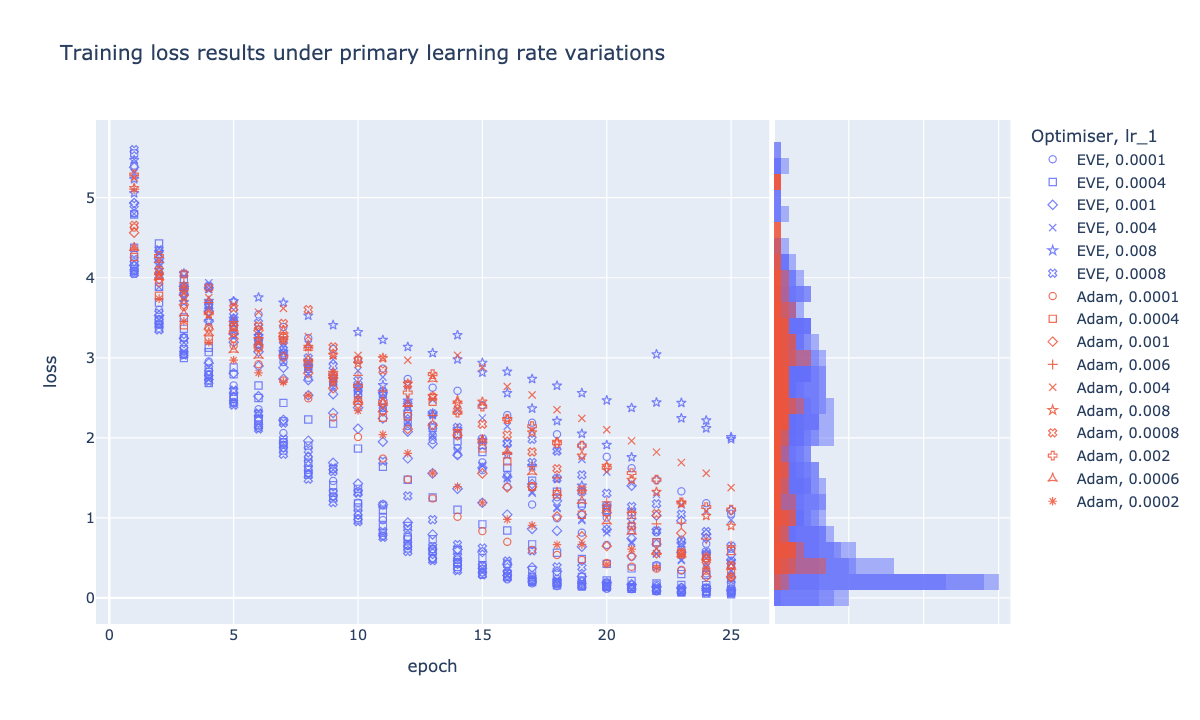}
    \caption{Illustration of the training loss under a range of primary learning rate variations. As shown, the marginal histogram of the training loss illustrates the superiority of EVE over Adam on CIFAR-100 data.}
    \label{fig:fig1}
\end{figure}

\begin{figure} 
    \centering
    \includegraphics[width=\textwidth]{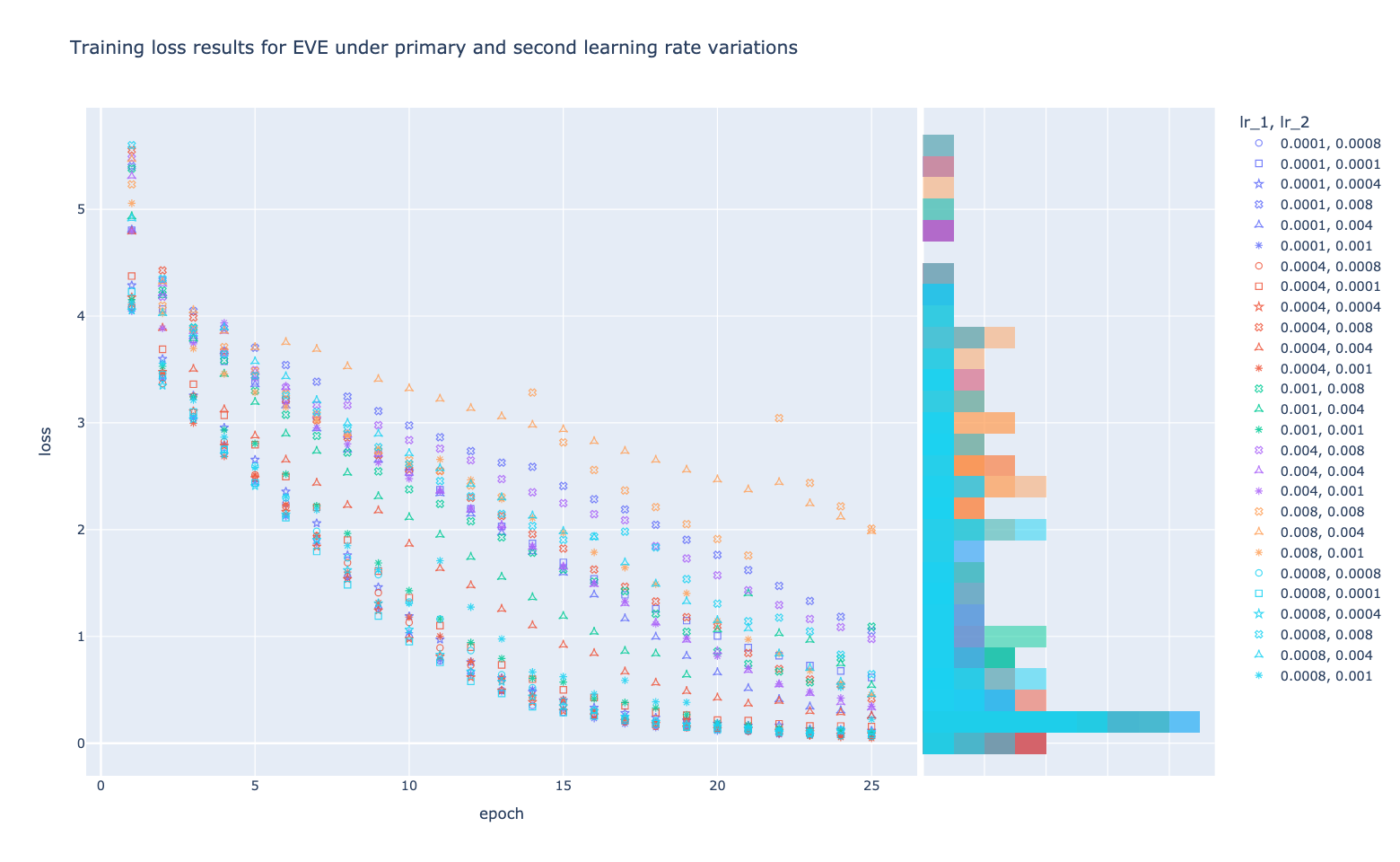}
    \caption{Illustration of the training loss under a range of primary and second learning rate variations for EVE. As shown, the various values for primary and second learning rate lead to the same performance according to the marginal training loss distribution with the majority of the settings hitting below 0.2999 as the training loss value.}
    \label{fig:fig2}
\end{figure}

Figures \ref{fig:fig1} and \ref{fig:fig2} illustrate the training loss under a variety of learning rates settings. We have visualised the trajectories of the training loss using primary learning rate variations for both Adam and EVE (Figure \ref{fig:fig1}). Moreover, Figure \ref{fig:fig2} shows the training loss trajectories for EVE using primary and secondary learning rate variations. The illustrations clearly demonstrate that EVE outperforms Adam. EVE consistently converges more quickly and attains lower training loss values across different learning rate adjustments, as supported by the training loss histogram.

From the results, it's evident that for lower or higher learning rates, the proposed optimisation method either outperforms or is competitive with Adam on the CIFAR-100 dataset using ResNet-50. However, as the learning rate increased to 0.01, Adam struggled with convergence, and its performance degraded. Yet, even in this scenario, the proposed method displayed a better performance over Adam.

The convergence epochs also shed light on the efficiency of the proposed method, converging faster than Adam in scenarios across various learning rates. This could be attributed to the dual and additional learning rates adapting better to the loss landscape of the model and dataset.

\begin{figure} 
    \centering
    \includegraphics[width=\textwidth]{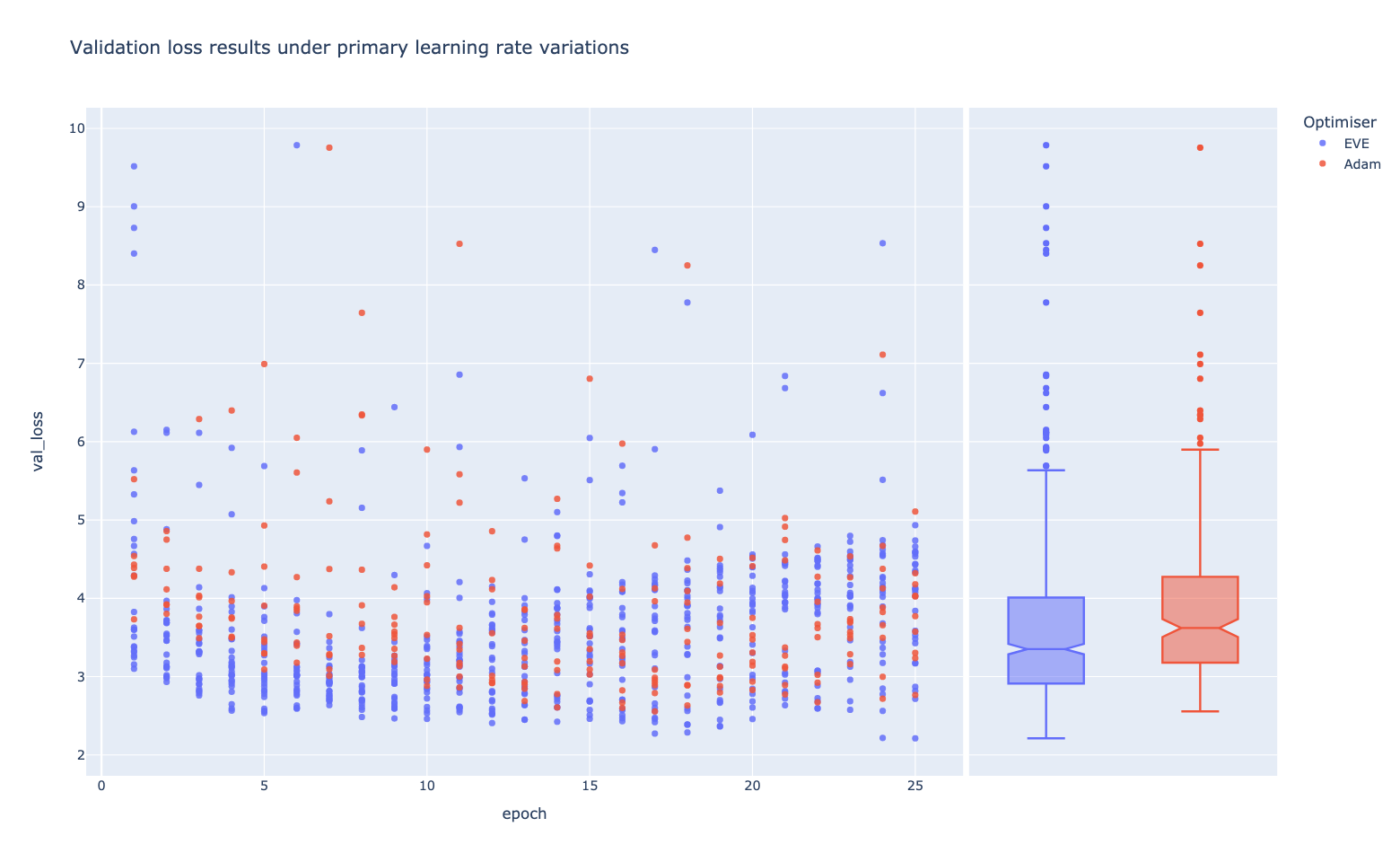}
    \caption{Illustration of the validation losses for EVE and Adam using CIFAR-100 dataset. As shown, EVE outperforms Adam by achieving consistent validation loss values during the training which will yield improved generalisation capacity.}
    \label{fig:fig3}
\end{figure}

Figure \ref{fig:fig3} illustrates the validation loss for various learning rate settings with Adam and EVE. Table \ref{table:table1} provides statistics related to the validation loss. As depicted, EVE surpasses Adam in terms of validation loss mean, median, standard deviation, and minimum validation loss. The significant disparity in the standard deviations of the validation loss suggests that EVE offers a more stable tuning experience compared to Adam, potentially resulting in enhanced generalisation.

\begin{table}
\centering
\begin{tabular}{ |c||c|c|c|c|  }
 \hline
 \multicolumn{5}{|c|}{Validation Loss Statistics} \\
 \hline
Optimisation & Median & Mean & STD & min\\
 \hline
 Adam   & 3.73    &251.93&   3,833.71&2.55\\
 EVE&   \textbf{3.37}  & \textbf{4.20}   &\textbf{7.18}&\textbf{2.21}\\
 \hline
\end{tabular}
\caption{Validation loss statistics for Adam and EVE using CIFAR-100 dataset and ResNet-50.}
\label{table:table1}
\end{table}

\subsection{Dataset: Flower Classification}
The dataset, available on Kaggle \footnote{\url{https://www.kaggle.com/competitions/flower-classification-with-tpus}}, encompasses images of 104 distinct flower species. It offers four different image resolutions, from 192x192 to 512x512 pixels. For our experiments, we selected the 224x224 pixel resolution to train the network. We divided the data into 14,609 training images and 1,856 validation samples, all representing the 104 flower varieties.

\subsubsection{Model Architecture: DenseNet-201}
DenseNet-201, a variant of the Dense Convolutional Network (DenseNet) architecture, stands out due to its unique approach to deep learning model design. At its core, DenseNet-201 emphasizes the creation of dense connections between layers, wherein each layer receives the feature-maps from all preceding layers and passes on its own to all subsequent layers. This fosters feature reuse, strengthens the gradient flow, and alleviates the vanishing gradient problem. Comprising 201 layers, DenseNet-201 offers a deeper architecture while maintaining computational efficiency, making it a favored choice for various computer vision tasks.

\subsubsection{Experimental Settings}
We used the following settings for our experiment using DenseNet-201 and Flower Classification data:
\begin{itemize}
    \item \textbf{Learning Rates}: The learning rates were varied between \{2e-5, 0.008\} for both EVE and Adam.
    \item \textbf{Batch Size}: Fixed at 128.
    \item \textbf{Epochs}: Models were trained for 25 epochs.
    \item \textbf{Hyper-parameters}: Set to default values.
    \item \textbf{Loss Function}: Categorical Focal Loss as described in \cite{khadangi2023deepflorist} was used.
    \item \textbf{Accelerated Computing}: Both Adam and EVE were trained using Google Colab TPU using 8 replicas.
\end{itemize}

\subsubsection{Evaluation Metric}
The primary metrics for evaluation were the training and validation loss, along with the Macro F1-score, which were used to study convergence properties and tendencies toward over-fitting.
\subsubsection{Results}
\begin{figure} 
    \centering
    \includegraphics[width=\textwidth]{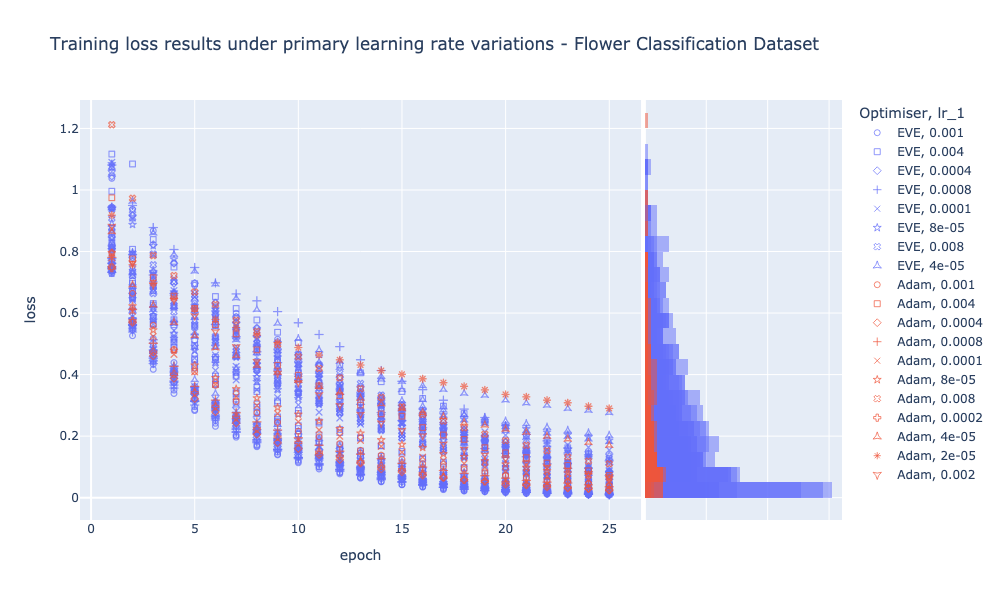}
    \caption{Illustration of the training loss under a range of primary learning rate variations. As shown, the marginal histogram of the training loss illustrates the superiority of EVE over Adam on Flower Classification Data.}
    \label{fig:fig4}
\end{figure}

\begin{figure} 
    \centering
    \includegraphics[width=\textwidth]{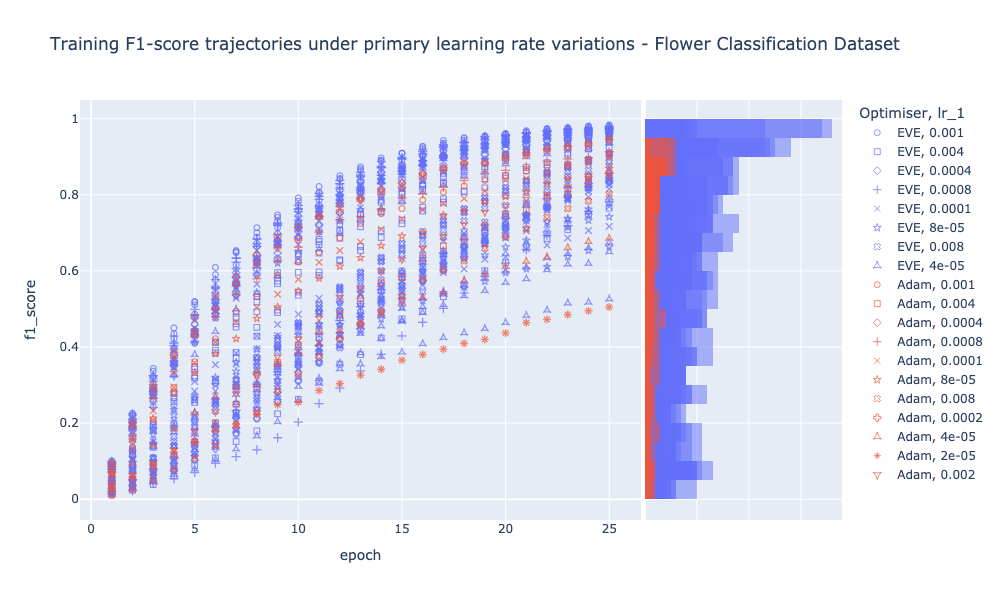}
    \caption{Illustration of the training F1-score trajectories under a range of primary learning rate variations. As shown, the marginal histogram of the training F1-score illustrates the superiority of EVE over Adam on Flower Classification data.}
    \label{fig:fig5}
\end{figure}

Figures \ref{fig:fig4} and \ref{fig:fig5} illustrate the training loss and F1-score under a variety of learning rates settings. We have visualised the trajectories of the training loss using primary learning rate variations for both Adam and EVE (Figure \ref{fig:fig4}). Moreover, Figure \ref{fig:fig5} shows the training F1-score trajectories for Adam and EVE using primary learning rate variations. The illustrations clearly demonstrate that EVE outperforms Adam. Across various learning rate settings, EVE consistently converges more quickly, attains a lower training loss, and achieves higher training F1-score values.

\begin{figure} 
    \centering
    \includegraphics[width=\textwidth]{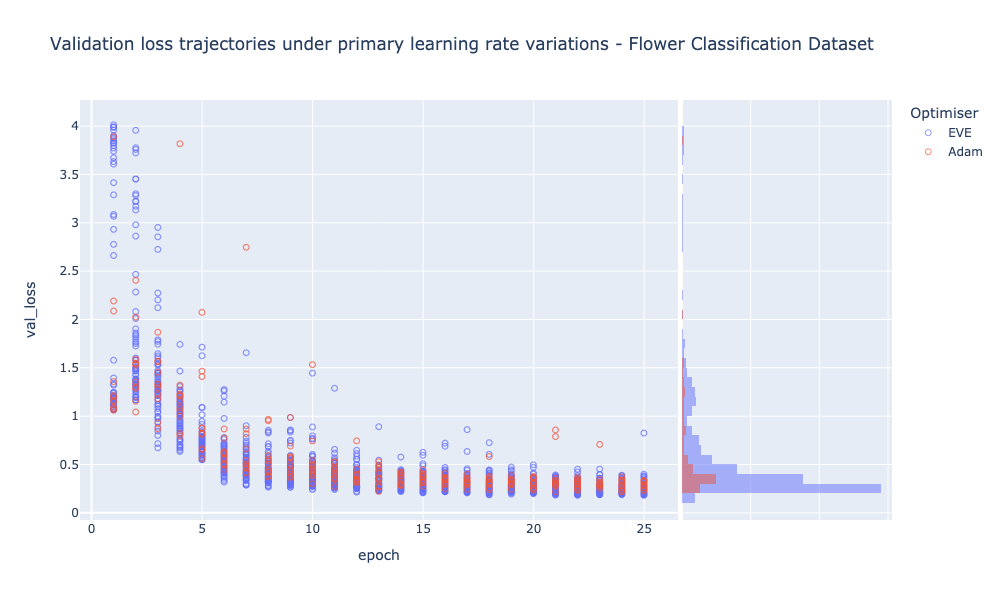}
    \caption{Illustration of the validation loss trajectories under a range of primary learning rate variations. As shown, the marginal histogram of the validation loss illustrates the superiority of EVE over Adam on Flower Classification data.}
    \label{fig:fig6}
\end{figure}

\begin{figure} 
    \centering
    \includegraphics[width=\textwidth]{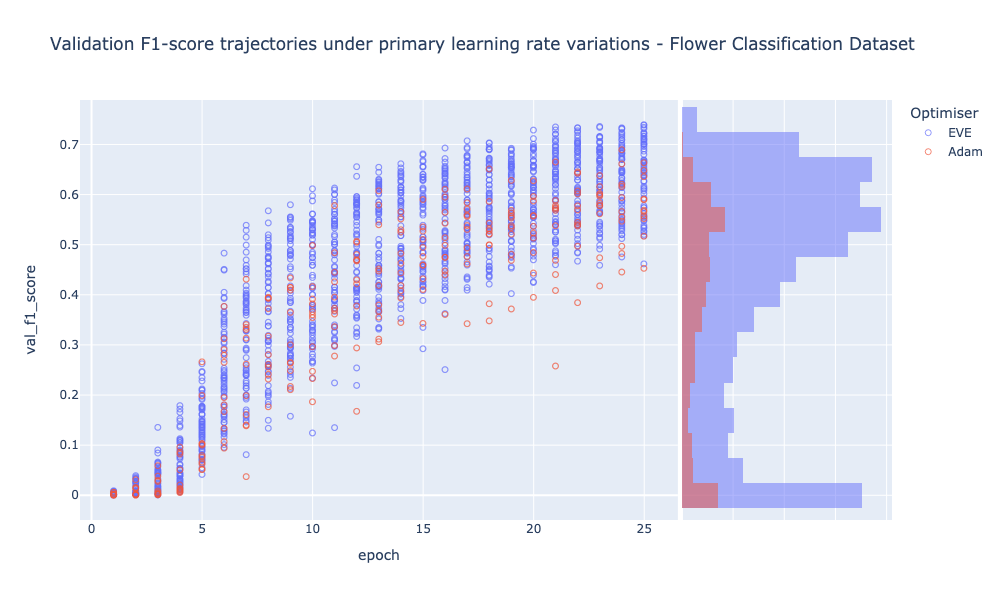}
    \caption{Illustration of the validation F1-score trajectories under a range of primary learning rate variations. As shown, the marginal histogram of the validation F1-score illustrates the superiority of EVE over Adam on Flower Classification data.}
    \label{fig:fig7}
\end{figure}

\begin{figure} 
    \centering
    \includegraphics[width=\textwidth]{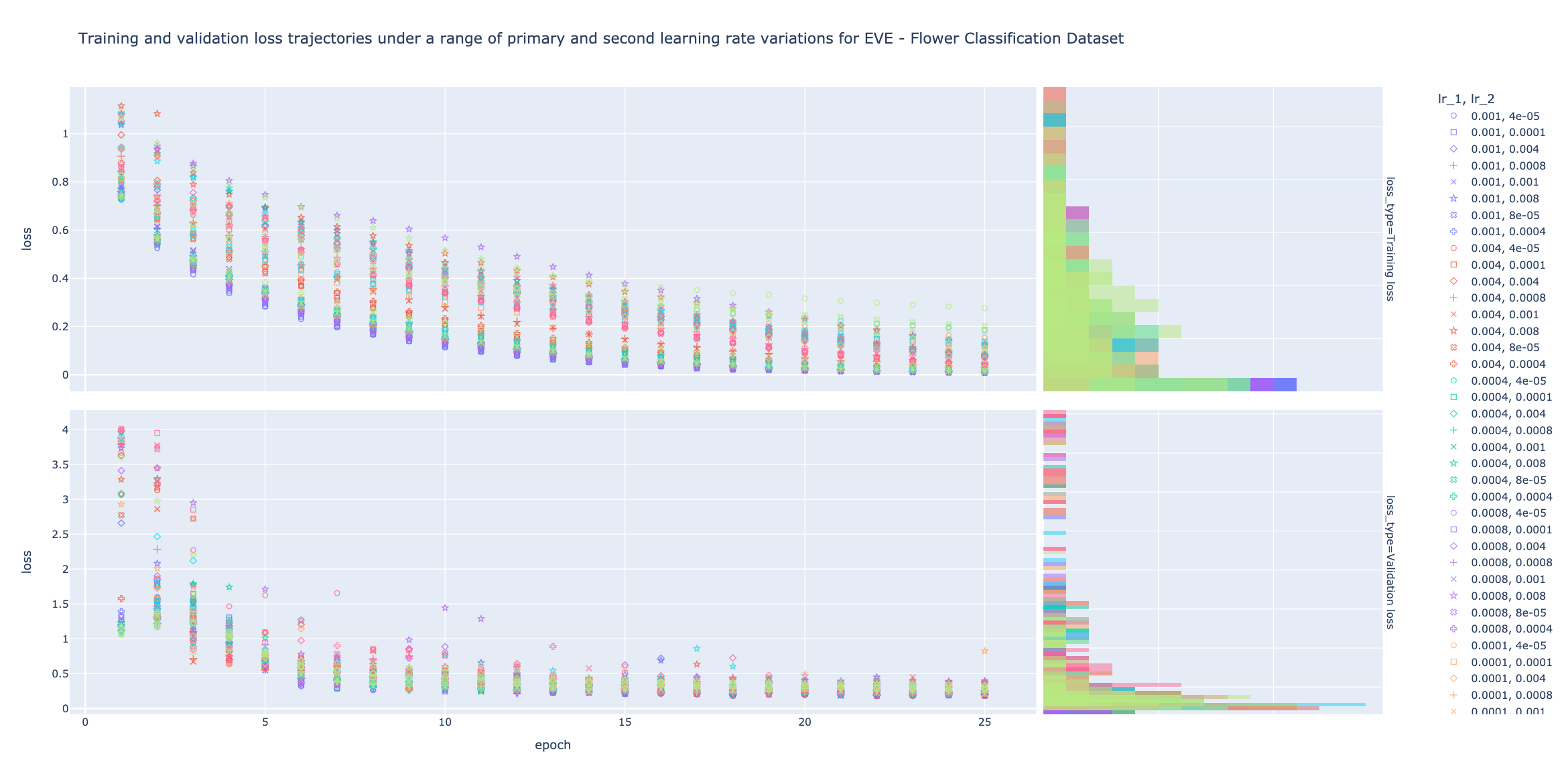}
    \caption{Illustration of the training and validation loss trajectories under a range of primary and second learning rate variations for EVE on Flower Classification data. As shown, the various values for primary and second learning rate lead to the same performance according to the marginal training and validation loss distributions with the majority of the settings hitting below 0.05 and 0.3 for the training and validation loss values, respectively.}
    \label{fig:fig8}
\end{figure}

\begin{figure} 
    \centering
    \includegraphics[width=\textwidth]{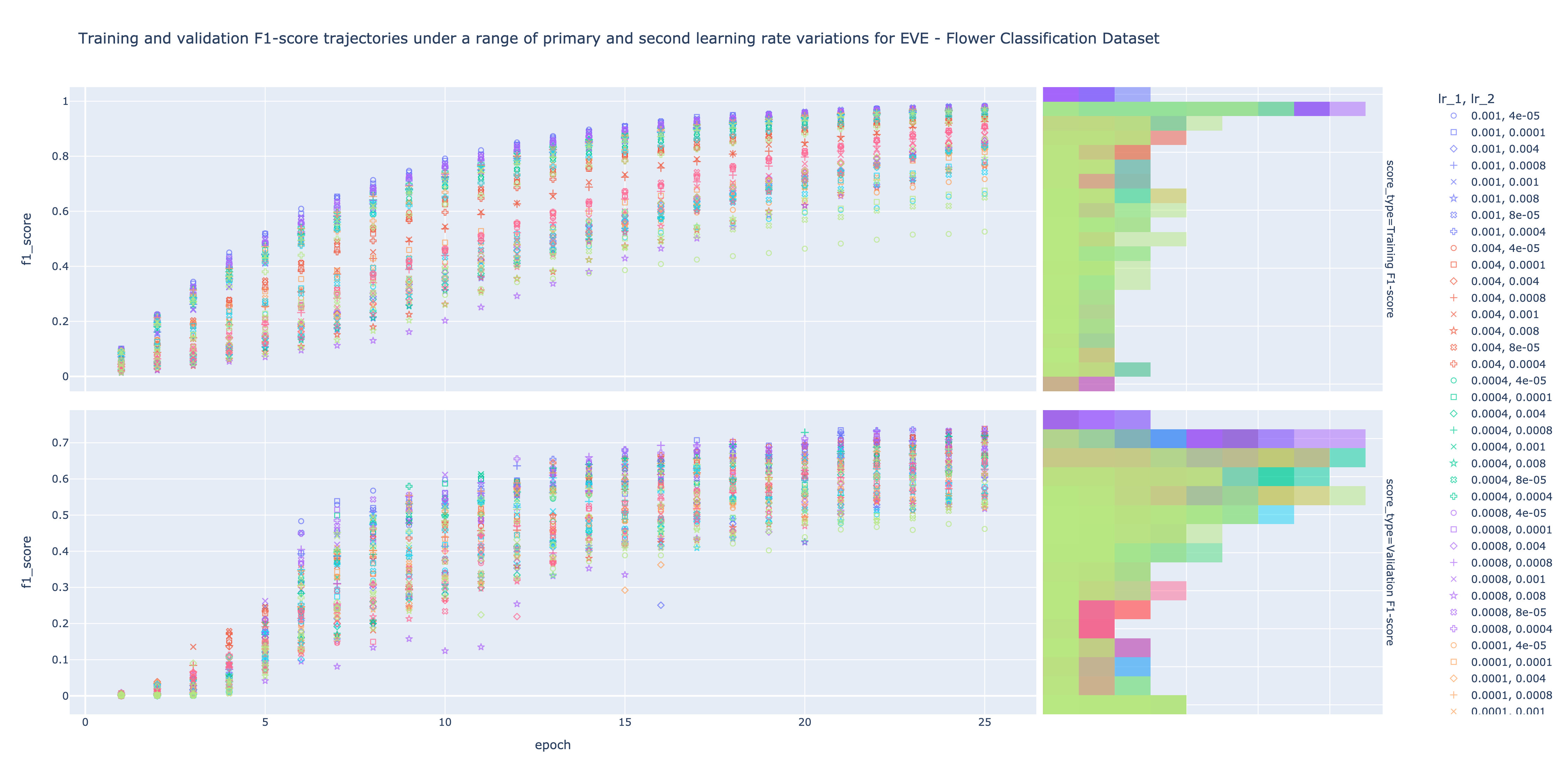}
    \caption{Illustration of the training and validation F1-score trajectories under a range of primary and second learning rate variations for EVE on Flower Classification data. As shown, the various values for primary and second learning rate lead to the same performance according to the marginal training and validation F1-score distributions with the majority of the settings.}
    \label{fig:fig9}
\end{figure}

Figures \ref{fig:fig6} and \ref{fig:fig7} illustrate the validation loss and F1-score under a variety of learning rates settings. Both illustrations show that, across various learning rate combinations, EVE consistently outperforms Adam in achieving better validation metrics during training. Finally, Figures \ref{fig:fig8} and \ref{fig:fig9} represent the training and validation loss and F1-score trajectories under primary and secondary learning rate variations for EVE. In Figure \ref{fig:fig8}, the top section displays the training loss, while the bottom section depicts the validation loss. Both trajectories indicate that EVE's performance remains consistent across various combinations of primary and secondary learning rates. In Figure \ref{fig:fig9}, the top section displays the training F1-score, while the bottom section depicts the validation F1-score. The trajectories show that EVE consistently performs well across a range of different primary and secondary learning rates.

\section{Discussion and Conclusion}
\subsection{Discussion}
In the arena of DNNs optimisation, the intricacy of choosing the right learning rate has always been a challenging task. Often, the learning rate is considered the most crucial hyper-parameter to adjust, dictating the convergence speed and the final model performance. This paper introduces a novel optimisation method that leverages dual learning rates along with dual momentum terms, addressing the potential limitations of traditional single learning rate-based methods.

One of the principal observations from our experiments is the advantage of having dual learning rates. While one learning rate is instrumental in guiding the general direction of optimisation, the second aids in fine-tuning, acting as a counterbalance, and offering more control over the learning process. 

Similarly, the introduction of dual momentum terms further refined the optimisation. Momentum, in essence, helps in accelerating the gradients vectors in the right directions, thus leading to faster convergence. By having two dual momentum terms, our model can learn from both its recent updates and long-term trends, providing a more holistic understanding of the gradient dynamics.

Comparing our proposed method with the widely-accepted Adam optimiser on two datasets - CIFAR-100 and Flower Classification - gave us fascinating insights. The consistency of our method's performance across different datasets and its consistency under various learning rate settings are particularly noteworthy.

\subsection{Conclusion}

The study concludes that optimisation methods with dual learning rates and dual momentum terms can significantly enhance the learning process. We showed that EVE consistently outperformed the Adam optimiser, particularly when tested on CIFAR-100 and Flower Classification datasets. The empirical results underscore the potential of our proposed method, marking a substantial step forward in the field.

Additionally, as we stride towards the world of more complex architectures and larger datasets, the need for more robust optimisation techniques becomes paramount. Our method is not just a testament to the viability of dual learning rates and momenta but also paves the way for more adaptive and dynamic optimisation strategies.

Looking forward, we see multiple avenues for improvement and further research:

\begin{itemize}
    \item Scalability: While the current results are promising, testing the optimisation method on larger datasets and more intricate architectures will be essential.
    \item Adaptability: Future iterations could explore the possibility of dynamically adjusting the learning rates and momentum values based on the training phase.
    \item Integration: Investigating how this optimisation method fares when integrated with other modern techniques, like adaptive gradient clipping or Lookahead method, could be an interesting direction.
\end{itemize}

Lastly, in the spirit of open science and to ensure reproducibility, we have provided the code for EVE on our GitHub \footnote{\url{https://github.com/akhadangi/EVE}} implemented using TensorFlow. We encourage the community to delve into it, offer feedback, and build upon our foundational work. Moreover, we have provided all the figures in interactive mode\footnote{\url{https://eve-optimiser.github.io}} to help the community explore the performance of EVE in more granular level.

\clearpage
\bibliographystyle{unsrt}  
\bibliography{references}

\begin{thebibliography}{10}

\bibitem{lecun2015deep}
Yann LeCun, Yoshua Bengio, and Geoffrey Hinton.
\newblock Deep learning.
\newblock {\em nature}, 521(7553):436--444, 2015.

\bibitem{goodfellow2016deep}
Ian Goodfellow, Yoshua Bengio, and Aaron Courville.
\newblock {\em Deep learning}.
\newblock MIT press, 2016.

\bibitem{bottou2010large}
L{\'e}on Bottou.
\newblock Large-scale machine learning with stochastic gradient descent.
\newblock In {\em Proceedings of COMPSTAT'2010: 19th International Conference
  on Computational StatisticsParis France, August 22-27, 2010 Keynote, Invited
  and Contributed Papers}, pages 177--186. Springer, 2010.

\bibitem{keskar2017improving}
Nitish~Shirish Keskar and Richard Socher.
\newblock Improving generalization performance by switching from adam to sgd.
\newblock {\em arXiv preprint arXiv:1712.07628}, 2017.

\bibitem{khadangi2021net}
Afshin Khadangi, Thomas Boudier, and Vijay Rajagopal.
\newblock Em-net: Deep learning for electron microscopy image segmentation.
\newblock In {\em 2020 25th international conference on pattern recognition
  (ICPR)}, pages 31--38. IEEE, 2021.

\bibitem{khadangi2021stellar}
Afshin Khadangi, Thomas Boudier, and Vijay Rajagopal.
\newblock Em-stellar: benchmarking deep learning for electron microscopy image
  segmentation.
\newblock {\em Bioinformatics}, 37(1):97--106, 2021.

\bibitem{khadangi2022cardiovinci}
Afshin Khadangi, Thomas Boudier, Eric Hanssen, and Vijay Rajagopal.
\newblock Cardiovinci: building blocks for virtual cardiac cells using deep
  learning.
\newblock {\em Philosophical Transactions of the Royal Society B},
  377(1864):20210469, 2022.

\bibitem{robbins1951stochastic}
Herbert Robbins and Sutton Monro.
\newblock A stochastic approximation method.
\newblock {\em The annals of mathematical statistics}, pages 400--407, 1951.

\bibitem{polyak1964some}
Boris~T Polyak.
\newblock Some methods of speeding up the convergence of iteration methods.
\newblock {\em Ussr computational mathematics and mathematical physics},
  4(5):1--17, 1964.

\bibitem{duchi2011adaptive}
John Duchi, Elad Hazan, and Yoram Singer.
\newblock Adaptive subgradient methods for online learning and stochastic
  optimization.
\newblock {\em Journal of machine learning research}, 12(7), 2011.

\bibitem{tieleman2012lecture}
Tijmen Tieleman, Geoffrey Hinton, et~al.
\newblock Lecture 6.5-rmsprop: Divide the gradient by a running average of its
  recent magnitude.
\newblock {\em COURSERA: Neural networks for machine learning}, 4(2):26--31,
  2012.

\bibitem{kingma2014adam}
Diederik~P Kingma and Jimmy Ba.
\newblock Adam: A method for stochastic optimization.
\newblock {\em arXiv preprint arXiv:1412.6980}, 2014.

\bibitem{bottou2018optimization}
L{\'e}on Bottou, Frank~E Curtis, and Jorge Nocedal.
\newblock Optimization methods for large-scale machine learning.
\newblock {\em SIAM review}, 60(2):223--311, 2018.

\bibitem{sutskever2013importance}
Ilya Sutskever, James Martens, George Dahl, and Geoffrey Hinton.
\newblock On the importance of initialization and momentum in deep learning.
\newblock In {\em International conference on machine learning}, pages
  1139--1147. PMLR, 2013.

\bibitem{schaul2013no}
Tom Schaul, Sixin Zhang, and Yann LeCun.
\newblock No more pesky learning rates.
\newblock In {\em International conference on machine learning}, pages
  343--351. PMLR, 2013.

\bibitem{li2017convergence}
Yuanzhi Li and Yang Yuan.
\newblock Convergence analysis of two-layer neural networks with relu
  activation.
\newblock {\em Advances in neural information processing systems}, 30, 2017.

\bibitem{glorot2010understanding}
Xavier Glorot and Yoshua Bengio.
\newblock Understanding the difficulty of training deep feedforward neural
  networks.
\newblock In {\em Proceedings of the thirteenth international conference on
  artificial intelligence and statistics}, pages 249--256. JMLR Workshop and
  Conference Proceedings, 2010.

\bibitem{pmlr-v38-choromanska15}
Anna Choromanska, MIkael Henaff, Michael Mathieu, Gerard Ben~Arous, and Yann
  LeCun.
\newblock {The Loss Surfaces of Multilayer Networks}.
\newblock In Guy Lebanon and S.~V.~N. Vishwanathan, editors, {\em Proceedings
  of the Eighteenth International Conference on Artificial Intelligence and
  Statistics}, volume~38 of {\em Proceedings of Machine Learning Research},
  pages 192--204, San Diego, California, USA, 09--12 May 2015. PMLR.

\bibitem{bergstra2012random}
James Bergstra and Yoshua Bengio.
\newblock Random search for hyper-parameter optimization.
\newblock {\em Journal of machine learning research}, 13(2), 2012.

\bibitem{dantzig1949programming}
George~B Dantzig.
\newblock Programming of interdependent activities: Ii mathematical model.
\newblock {\em Econometrica, Journal of the Econometric Society}, pages
  200--211, 1949.

\bibitem{khachiyan1979polynomial}
Leonid~G Khachiyan.
\newblock A polynomial algorithm in linear programming (english translation).
\newblock In {\em Soviet Mathematics Doklady}, volume~20, pages 191--194, 1979.

\bibitem{karmarkar1984new}
Narendra Karmarkar.
\newblock A new polynomial-time algorithm for linear programming.
\newblock In {\em Proceedings of the sixteenth annual ACM symposium on Theory
  of computing}, pages 302--311, 1984.

\bibitem{wright1997primal}
Stephen~J Wright.
\newblock {\em Primal-dual interior-point methods}.
\newblock SIAM, 1997.

\bibitem{mehrotra1992implementation}
Sanjay Mehrotra.
\newblock On the implementation of a primal-dual interior point method.
\newblock {\em SIAM Journal on optimization}, 2(4):575--601, 1992.

\bibitem{cauchy1847methode}
Augustin Cauchy et~al.
\newblock M{\'e}thode g{\'e}n{\'e}rale pour la r{\'e}solution des systemes
  d’{\'e}quations simultan{\'e}es.
\newblock {\em Comp. Rend. Sci. Paris}, 25(1847):536--538, 1847.

\bibitem{hestenes1952methods}
Magnus~R Hestenes, Eduard Stiefel, et~al.
\newblock Methods of conjugate gradients for solving linear systems.
\newblock {\em Journal of research of the National Bureau of Standards},
  49(6):409--436, 1952.

\bibitem{broyden1967quasi}
Charles~G Broyden.
\newblock Quasi-newton methods and their application to function minimisation.
\newblock {\em Mathematics of Computation}, 21(99):368--381, 1967.

\bibitem{bellman1957markovian}
Richard Bellman.
\newblock A markovian decision process.
\newblock {\em Journal of mathematics and mechanics}, pages 679--684, 1957.

\bibitem{pareto1896curva}
Vilfredo Pareto.
\newblock La curva delle entrate e le osservazioni del prof. edgeworth.
\newblock {\em Giornale degli economisti}, pages 439--448, 1896.

\bibitem{zeleny1973compromise}
Milan Zeleny.
\newblock Compromise programming.
\newblock {\em Multiple criteria decision making}, 1973.

\bibitem{kirkpatrick1983optimization}
Scott Kirkpatrick, C~Daniel Gelatt~Jr, and Mario~P Vecchi.
\newblock Optimization by simulated annealing.
\newblock {\em science}, 220(4598):671--680, 1983.

\bibitem{holland1992adaptation}
John~H Holland.
\newblock {\em Adaptation in natural and artificial systems: an introductory
  analysis with applications to biology, control, and artificial intelligence}.
\newblock MIT press, 1992.

\bibitem{boyd2004convex}
Stephen~P Boyd and Lieven Vandenberghe.
\newblock {\em Convex optimization}.
\newblock Cambridge university press, 2004.

\bibitem{kuhn1955hungarian}
Harold~W Kuhn.
\newblock The hungarian method for the assignment problem.
\newblock {\em Naval research logistics quarterly}, 2(1-2):83--97, 1955.

\bibitem{ford1956maximal}
Lester~Randolph Ford and Delbert~R Fulkerson.
\newblock Maximal flow through a network.
\newblock {\em Canadian journal of Mathematics}, 8:399--404, 1956.

\bibitem{nesterov1983method}
Yurii~Evgen'evich Nesterov.
\newblock A method of solving a convex programming problem with convergence
  rate o$\backslash$bigl(k\^{}2$\backslash$bigr).
\newblock In {\em Doklady Akademii Nauk}, volume 269, pages 543--547. Russian
  Academy of Sciences, 1983.

\bibitem{loshchilov2017decoupled}
Ilya Loshchilov and Frank Hutter.
\newblock Decoupled weight decay regularization.
\newblock {\em arXiv preprint arXiv:1711.05101}, 2017.

\bibitem{reddi2019convergence}
Sashank~J Reddi, Satyen Kale, and Sanjiv Kumar.
\newblock On the convergence of adam and beyond.
\newblock {\em arXiv preprint arXiv:1904.09237}, 2019.

\bibitem{bengio2013advances}
Yoshua Bengio, Nicolas Boulanger-Lewandowski, and Razvan Pascanu.
\newblock Advances in optimizing recurrent networks.
\newblock In {\em 2013 IEEE international conference on acoustics, speech and
  signal processing}, pages 8624--8628. IEEE, 2013.

\bibitem{liner2021improving}
Elizabeth Liner and Risto Miikkulainen.
\newblock Improving neural network learning through dual variable learning
  rates.
\newblock In {\em 2021 International Joint Conference on Neural Networks
  (IJCNN)}, pages 1--7. IEEE, 2021.

\bibitem{sun2023adasam}
Hao Sun, Li~Shen, Qihuang Zhong, Liang Ding, Shixiang Chen, Jingwei Sun, Jing
  Li, Guangzhong Sun, and Dacheng Tao.
\newblock Adasam: Boosting sharpness-aware minimization with adaptive learning
  rate and momentum for training deep neural networks.
\newblock {\em arXiv preprint arXiv:2303.00565}, 2023.

\bibitem{krizhevsky2009learning}
Alex Krizhevsky, Geoffrey Hinton, et~al.
\newblock Learning multiple layers of features from tiny images.
\newblock {\em University of Toronto}, 2009.

\bibitem{he2016deep}
Kaiming He, Xiangyu Zhang, Shaoqing Ren, and Jian Sun.
\newblock Deep residual learning for image recognition.
\newblock In {\em Proceedings of the IEEE conference on computer vision and
  pattern recognition}, pages 770--778, 2016.

\bibitem{khadangi2023deepflorist}
Afshin Khadangi.
\newblock Deepflorist: Rethinking deep neural networks and ensemble learning as
  a meta-classifier for object classification.
\newblock {\em arXiv preprint arXiv:2307.01806}, 2023.

\bibitem{huang2017densely}
Gao Huang, Zhuang Liu, Laurens Van Der~Maaten, and Kilian~Q Weinberger.
\newblock Densely connected convolutional networks.
\newblock In {\em Proceedings of the IEEE conference on computer vision and
  pattern recognition}, pages 4700--4708, 2017.

\end{thebibliography}


\end{document}